\documentclass[letterpaper]{article}
\usepackage{iccc}

\usepackage{times}
\usepackage{helvet}
\usepackage{courier}
\usepackage{amsmath, amssymb}
\usepackage{graphicx}
\usepackage{amsfonts}
\usepackage{dsfont}
\usepackage{tabularx}
\usepackage{multirow}
\usepackage{algorithm}
\usepackage{algpseudocode}
\usepackage{float}
\usepackage{multicol}
\usepackage[title]{appendix}

\usepackage{cuted}
\usepackage{capt-of}

\title{ORACLE: Leveraging Mutual Information for C\underline{o}nsistent Cha\underline{rac}ter Generation with \underline{L}oRAs in Diffusion Mod\underline{e}ls}
\author{Kiymet Akdemir, Pinar Yanardag \\
Department of Computer Science, Virginia Tech \\
Blacksburg, Virginia, USA \\
\{kiymet, pinary\}@vt.edu 
}
\begin{document} 
\maketitle


\begin{strip}\centering
\vspace{-1cm}
\includegraphics[width=\textwidth]{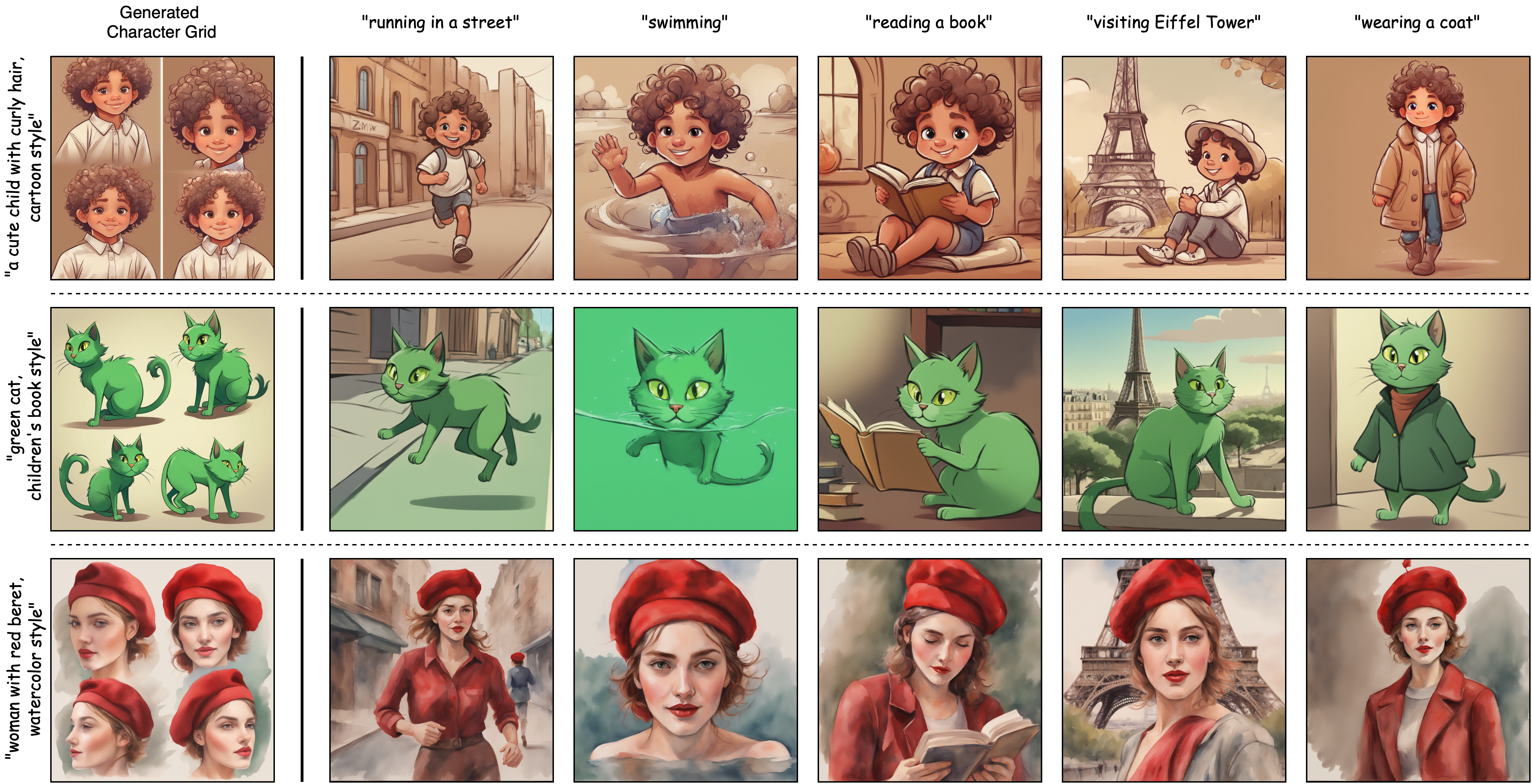}
\captionof{figure}{Given a text prompt such as `a cute child with curly chair, cartoon style' (refer to the top row), our approach seamlessly produces consistent characters in a zero-shot manner by leveraging a pre-trained Stable Diffusion model. It ensures character consistency across a wide array of settings and backgrounds, demonstrating the versatility and practicality of our method.  Our method has the potential to enhance creative process in art and design, enabling more detailed storytelling and consistent character portrayal in animations, video games, and interactive media.
\label{fig:teaser}}
\end{strip}

\begin{abstract}
\begin{quote}
Text-to-image diffusion models have recently taken center stage as pivotal tools in promoting visual creativity across an array of domains such as comic book artistry, children's literature, game development, and web design. These models harness the power of artificial intelligence to convert textual descriptions into vivid images, thereby enabling artists and creators to bring their imaginative concepts to life with unprecedented ease. However, one of the significant hurdles that persist is the challenge of maintaining consistency in character generation across diverse contexts. Variations in textual prompts, even if minor, can yield vastly different visual outputs, posing a considerable problem in projects that require a uniform representation of characters throughout. In this paper, we introduce a novel framework designed to produce consistent character representations from a single text prompt across diverse settings.  Through both quantitative and qualitative analyses, we demonstrate that our framework outperforms existing methods in generating characters with consistent visual identities, underscoring its potential to transform creative industries. By addressing the critical challenge of character consistency, we not only enhance the practical utility of these models but also broaden the horizons for artistic and creative expression. 

\end{quote}
\end{abstract}

\section{Introduction}
\label{sec:intro}

Text-to-image diffusion models have captivated the creative world with their extraordinary capacity to turn textual descriptions into detailed, high-resolution images. These advancements have ushered in a new era of creativity, allowing for the generation of bespoke illustrations for storybooks, dynamic characters in video games, personalized content across digital platforms, and engaging visuals for educational purposes. The ability to generate images that closely align with specific text prompts has opened up endless possibilities for storytellers, educators, game developers, and digital content creators, enabling them to bring their unique visions to life with precision and flair.

However, the journey of integrating these models into creative workflows has encountered a significant challenge: maintaining visual consistency across different scenarios. When characters are depicted in various contexts or settings, slight alterations in text prompts can lead to inconsistencies in their appearance, disrupting the visual continuity that is crucial for storytelling, brand identity, and character recognition. This challenge has been a bottleneck, limiting the full exploitation of text-to-image diffusion models in projects requiring a cohesive character narrative. 

In addressing the challenge of achieving consistent character visualization across various applications of text-to-image diffusion models, the field has increasingly leaned on personalization techniques such as Dreambooth \cite{ruiz2022dreambooth}, Textual Inversion~\cite{gal2022textual} or LoRAs \cite{hu2021lora}. Historically, these approaches have relied extensively on reference images for character creation, a dependency that constrains their applicability across a wider range of uses. Efforts to bypass these limitations have included strategies such as manual filtering and clustering~\cite{avrahami2023chosen}, or even the incorporation of celebrity names into prompts to guide the image synthesis process. However, such methods are typically either labor-intensive, time-consuming, or significantly restrict the diversity of characters that can be effectively rendered.

Our paper addresses this pivotal issue by presenting a novel approach that ensures the consistent generation of characters across diverse settings with a single text prompt. Based on a text prompt, for example, \textit{"a cute child with curly hair, cartoon style"} (refer to Figure~\ref{fig:teaser}), our method produces a set of initial character images in a zero-shot fashion through a pre-trained text-to-image diffusion model like Stable Diffusion \cite{rombach2022high}. This candidate set undergoes refinement through a mutual information-based filtering process which then serves as the foundation for training a personalization model, such as LoRA \cite{hu2021lora}. Following this process, it becomes possible to create characters that maintain consistency in a variety of settings, including diverse environments, backgrounds, and actions.

By enhancing the ability of diffusion models to maintain visual continuity, our methodology not only solves a technical problem but also profoundly impacts the creative process across multiple domains. For comic book artists and children's book authors, this breakthrough means characters can now retain their identity across panels or pages without the exhaustive effort of manual adjustments or the need for numerous reference images. This consistency is vital for narrative coherence and character development, allowing creators to focus on storytelling rather than technical limitations. In the realm of game development, our approach enables designers to create more immersive worlds, with characters that remain true to their original design throughout various game environments and scenarios. This consistency enhances the player's connection to the character and the overall gaming experience, allowing for a deeper engagement with the story. For educators and creators of educational content, this technology offers the potential to produce a wide range of consistent visual materials that can support learning objectives. Characters that recur in various educational scenarios can become memorable figures for students, aiding in engagement and the retention of information.

Our contributions are as follows:

\begin{itemize}
\item We propose an effective framework for producing characters that remain visually consistent across various scenarios. Our method operates in a zero-shot manner, generating unique characters that match the provided text prompts. It also eliminates discrepancies among image components using mutual information, ensuring cohesive visual representations by refining the initial set of generated images.

\item We provide comprehensive qualitative and quantitative comparisons with existing methods, along with insights from a user study, highlighting the effectiveness and improvements our method provides over traditional techniques.

\item The versatility and applicability of our method are highlighted through demonstrations of its use in various creative and practical contexts. We showcase how our approach enables the generation of characters that are not only consistent in appearance but also adaptable to different environments, backgrounds, and narratives, thereby broadening the potential for innovative applications in storytelling, gaming, education, and beyond.

\item Additionally, we illustrate how our approach can be utilized to design compelling storylines with a story example, and transform our characters into 3D objects for gaming purposes. This demonstrates our method's ability to not only create visually consistent characters but also to support the broader creative processes involved in narrative development and interactive game design. 

\end{itemize}

These contributions collectively enhance the toolbox available to creators and developers, offering new pathways to leverage text-to-image diffusion models for creating coherent and engaging visual narratives.

\section{Related Work}
\label{sec:related}

\subsection{Text-to-image Generation}
The advancement of large-scale text-to-image diffusion models~\cite{rombach2022high,nichol2021glide} has enabled the widespread use of image generation models~\cite{chen2020uniter,kim2021vilt,saharia2022photorealistic}, largely due to the plentiful availability of image-text pair datasets and their simpler training process when compared to Generative Adversarial Networks (GANs)~\cite{goodfellow2014generative}. While these models excel in producing a wide range of realistic images, they struggle to generate consistent images; minor changes in the prompt can lead to substantial variations in the outputs. This limitation restricts their applicability for illustration purposes, where consistency is key.

\subsection{Text-to-image Personalization}
Current text-to-image models struggle to depict specific entities across varied contexts. However, advancements in personalization techniques allow for the creation of images depicting a given subject in various contexts using a reference image set. Textual Inversion~\cite{gal2022textual} aims to integrate particular instances or styles into the embedding space of existing, unmodified text-to-image models. DreamBooth~\cite{ruiz2022dreambooth}, on the other hand, suggests fine-tuning the entire model to associate specific instances with a unique identifier, while still maintaining the instance's general category. This approach, though, necessitates saving a separate model for each subject, leading to storage issues. LoRA~\cite{hu2021lora} enables the fine-tuning of a limited number of parameters, significantly simplifying the storage of numerous personalized models. IP Adapter~\cite{ye2023ip-adapter}, employs an image encoder to integrate image features into the diffusion model. Nevertheless, it struggles accurately adhering to text prompts across varying contexts. These techniques are limited by the need for a reference image set provided by the user, which constrains the diversity and creativity of the subjects illustrated.

\subsection{Generating Consistent Characters}
Recent attempts in story illustration face limitations, as they are trained on specialized datasets~\cite{rahman2023makeastory}, depend on personalization models that require a set of reference images~\cite{gong2023talecrafter}, or utilize face-swapping techniques~\cite{jeong2023zeroshot}, which confines their use of human subjects. Consequently, these story illustration methods often limit character representations to pre-existing entities. Alternative approaches involve either manual image filtering or incorporating celebrity names into prompts, with the former being time-consuming and the latter narrowly confining the range of potential subjects for illustrations. The Chosen One~\cite{avrahami2023chosen} touches on the problem of generating imaginative characters and suggests generating and clustering images based on a specific prompt, then using the most consistent image cluster to iteratively train a personalized model until it reaches convergence. Yet, it demands considerable time for image generation and multiple rounds of training. 
Consistory~\cite{tewel2024trainingfree} and Story Diffusion~\cite{Zhou2024storydiffusion} use attention maps to ensure consistency across images.

\section{Background}
\label{sec:background}

\subsection{Diffusion models} Diffusion models are a class of generative models that estimates the complex data distribution through iterative denoising process. As the source of diversity, the initial latent $x_T \sim \mathcal{N}(\textbf{0}, \textbf{I})$ is sampled and fed to U-Net model, $\epsilon_\theta$, as an attempt to gradually denoise the latent variable $x_t$ to obtain $x_{t-1}$ for \textit{T} timesteps where $x_1$ corresponds to the final image. The joint probability of latent variables $\{x_1, ..., x_T\}$ is modeled as a Markov Chain.
\begin{equation}
    p_\theta(x_{1:T}) = p(x_T)\prod_{t=T}^{1}p_\theta(x_{t-1}|x_t)
\end{equation}
In text-to-image generation task, diffusion models are conditioned on an external text input $c$ where the overall aim is producing an image that is aligned with the description provided by $c$. To train text-to-image models, diffusion models use a simplified objective.
\begin{equation}
    \mathbb{E}_{x, c, \epsilon, t}\left[\lVert \epsilon_\theta(x_t, t, c) - \epsilon \rVert_2^2\right],
\end{equation}
where $(x_t, c)$ is latent-text condition pair, $\epsilon \sim \mathcal{N}(\textbf{0}, \textbf{I})$ and $t \sim \mathcal{U}([0, 1])$. For unconditional generation, $c$ is set to null text. In inference stage, classifier free guidance is applied to noise prediction to improve the sample quality.
\begin{equation}
    \Tilde{\epsilon}_\theta(x_t, t, c) = \epsilon_\theta(x_t, t, \emptyset) + \gamma[\epsilon_\theta(x_t, t, c) - \epsilon_\theta(x_t, t, \emptyset)],
\end{equation}
where $\gamma \geq 1$ is the guidance scale and $\emptyset$ denotes the null text condition.

\begin{figure*}[ht]
    \centering
    \includegraphics[width=\textwidth]{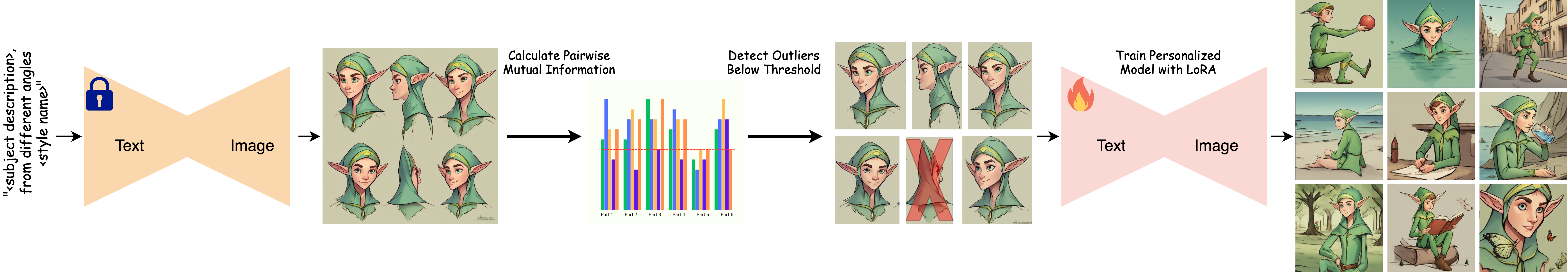}
    \caption{\textbf{An overview of ORACLE.} Our method operates through three phases: 1) It begins with the generation of a grid based on structured prompts that include character description, style, and a grid generator prompt, like \textit{"from different angles"}. 2) Subsequently, it calculates the average pairwise mutual information to identify potential outliers. 3) Once outliers are filtered out, a personalized model is trained using the refined grid segments. }
    \label{fig:framework}
\end{figure*}

\subsection{Mutual Information}
Using mutual information in the field of computer vision, first proposed by~\cite{viola1997alignment,maes1997multimodality} and has been employed as a robust method for comparing image similarity~\cite{4374125,kothandaraman2023aerialbooth} by binning pixel values into histograms and comparing their distributions.  Mutual information serves as a metric to measure the information acquired about one variable upon observing another variable. It effectively captures the dependence between two random variables, X and Y, using the following equation:
\begin{equation}
    \begin{split}
    I(X;Y)=H(X)-H(X|Y) \\
    =H(Y)-H(Y|X) \\
    =H(X)+H(Y)-H(X,Y)
    \end{split}
    \label{eq:mi}
\end{equation}

where 

\begin{equation}
    H(X)=-\sum_x P_X(x) \log P_X(x) = - E_{P_X} \log P_X,
\end{equation}

\begin{equation}
\begin{split}
    H(X|Y)=\sum_y P_Y(y) \left[ - \sum_x P_{X|Y}(x|y) \log
\left(P_{X|Y}(x|y)\right)\right] \\
= E_{P_Y} \left[ - E_{P_{X|Y}} \log P_{X|Y} \right]
\end{split}
\end{equation}

Entropy, a core concept in information theory, quantifies the level of unpredictability associated with a variable's outcomes. Here, $H(X)$ represents the entropy of X, signifying the inherent uncertainty or randomness of X, while $H(X|Y)$ denotes the conditional entropy of X given Y, which measures the remaining uncertainty in X once Y is known. The conditional entropy, $H(X|Y)$, specifically quantifies the extent to which the uncertainty of X is reduced by knowing the outcome of Y. The mutual information formula, therefore, quantifies the reduction in uncertainty about one variable given knowledge of the other.




\section{Method}
\label{sec:method}

Our approach is structured around three key phases. The initial part involves generating a grid of images that align with the provided prompt, using the text-to-image diffusion model. The second stage focuses on identifying and removing any outliers that do not match the consistency of the rest of the images. Lastly, we tailor a model using personalization techniques specifically designed to generate images across various contexts while maintaining the details in the set of refined consistent images. An overview of our framework is given in Figure~\ref{fig:framework}.

\begin{figure*}[]
    \centering
    \includegraphics[width=\textwidth]{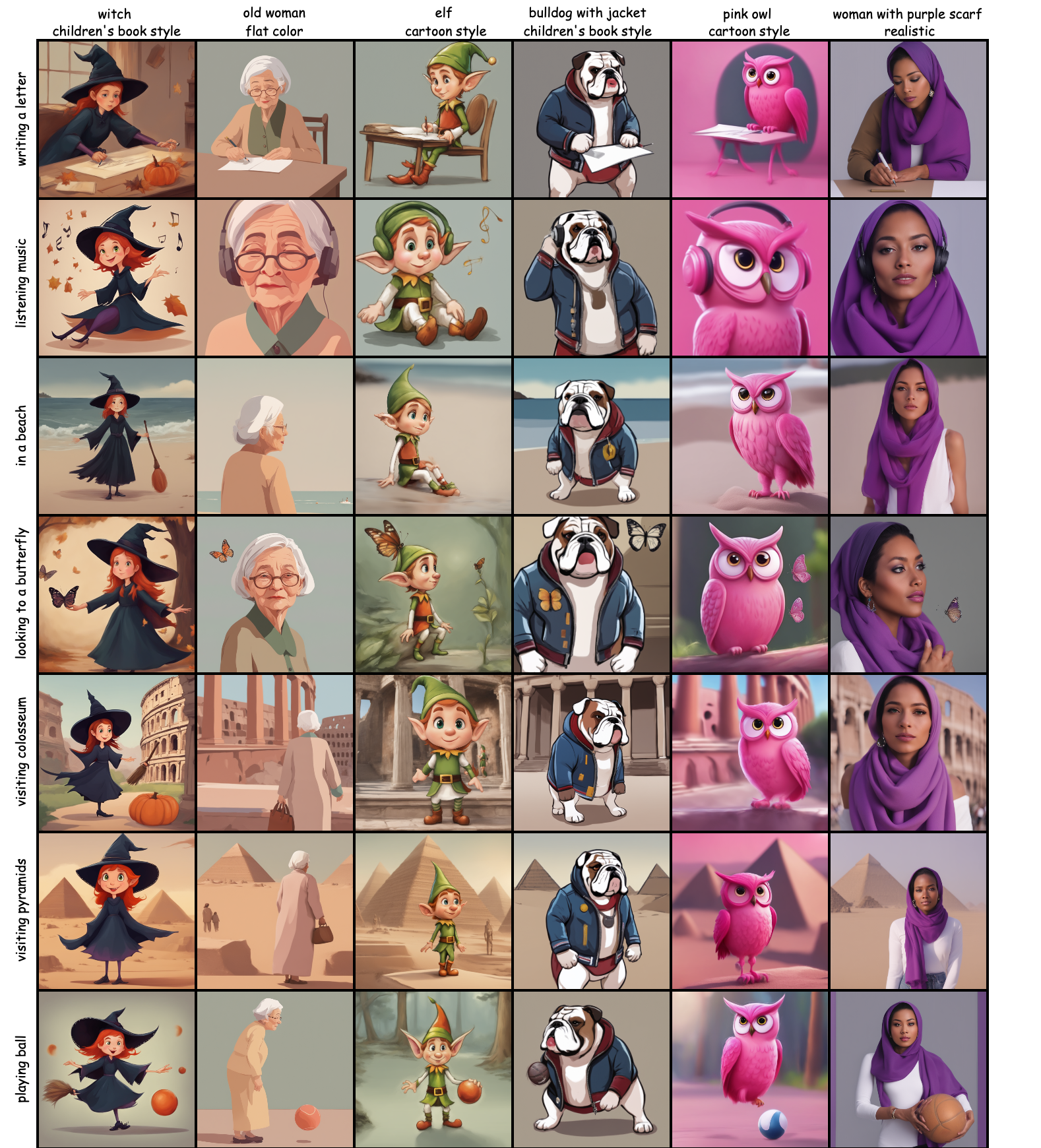}
    \caption{\textbf{Qualitative results.} Our method can produce a wide array of characters in diverse contexts and styles, from imaginative figures like \textit{`a bulldog wearing a jacket'} and \textit{`a pink owl'}, to photo-realistic characters such as \textit{`a woman with a purple scarf'}.}
    \label{fig:qual-result}
\end{figure*}
\subsection{Candidate Set Generation}
Given a text prompt, the initial step in our method involves generating a single grid of candidate images that correspond to the described character. In contrast to traditional approaches that depend on costly techniques such as clustering a large number of images \cite{avrahami2023chosen}, or the need for manually curated datasets to train a personalized model \cite{ruiz2022dreambooth,gal2022textual,hu2021lora}, our technique employs the "grid trick" to generate an initial set of candidates. This strategy, also referred to as a \textit{character sheet}, has become popular within the Stable Diffusion  art community, especially among enthusiasts and professionals for tasks like avatar creation and image stylization\footnote{How To Create Consistent Characters In Midjourney
: https://shorturl.at/jwAJW}. The trick involves leveraging a pre-trained text-to-image model with specific directions, such as \textit{"$<$character description$>$ from multiple angles, $<$style description$>$"} or \textit{"$<$character description$>$ from different perspectives, $<$style description$>$"}. While a template grid combined with ControlNet can be used to automatically crop image parts, we found that using a template compromises from the quality and creativity of the generated characters. Therefore, we prefer to manually crop the image parts, typically involving only 4-6 sections in a character grid. Previous research such as \cite{kara2023rave} have applied this technique for video editing and highlighted its ability to generate multiple images with a consistent style is due to the diffusion model treating the grid as a singular, composite image. 

\begin{figure*}
\begin{multicols}{2}
    \centering
    \includegraphics[width=0.5\textwidth]{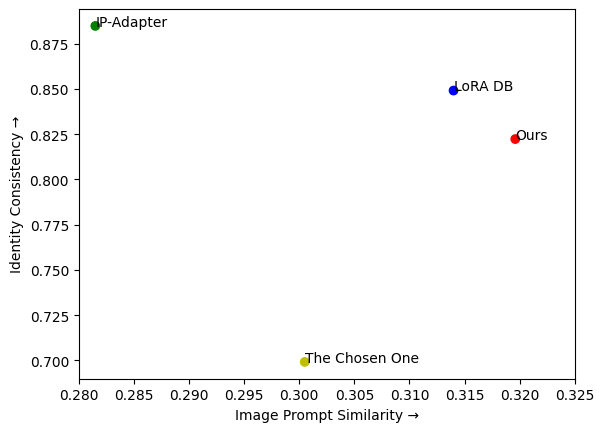}
    \caption{\textbf{Quantitative comparisons.} We use CLIP to assess the relevance of images to their prompts (image-prompt similarity) and identity consistency (image-image similarity).}
    \label{fig:quant-result}\par 
    \centering
    \includegraphics[width=0.48\textwidth]{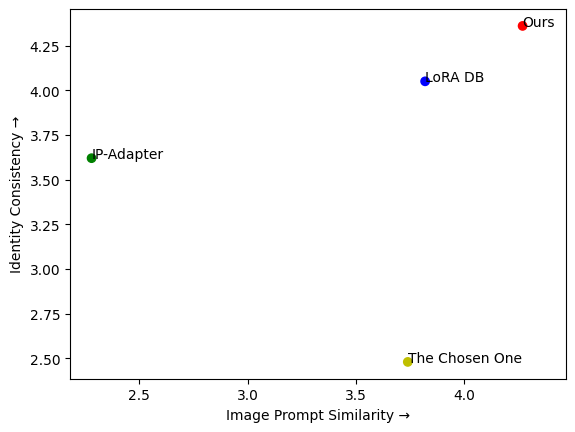}
    \caption{\textbf{User study results.} The average user rating for each baseline is given for two types of questions (identity consistency and relevance to prompt). Rating is performed on a scale from 1 to 5.}
    \label{fig:user-study}
    \end{multicols}
\end{figure*}

\subsection{Candidate Set Refinement}

While the initial batch of images accurately reflects the text prompt and achieves a level of consistency, it also displays noticeable inconsistencies, such as imprecise details or significant variations (illustrated by the leftmost set of images in Fig. \ref{fig:framework}). Therefore, we employ a mutual information-based strategy to identify and remove elements that could disrupt the uniformity of the personalized model. Examples are provided in Figure \ref{fig:appendix}. We argue that traditional vector similarity metrics, such as cosine similarity, fall short of our needs because they tend to interpret different views of the same subject as distinct features. In contrast, mutual information proves to be well-suited for our objective by assessing the distribution of image features, offering a more nuanced and effective means of evaluating consistency across various representations. 

Our objective is to generate a consistent set of images such that the average pairwise mutual information for each image within the set, i.e. $S_i = \frac{1}{k} \sum_{j=1}^k I(V_i, V_j)$ surpasses the predefined threshold. To achieve this, we have a binary function $\mathbf{C}$ that determines whether a specific image $V_i$ qualifies to be part of the final collection or not:
\begin{equation}
    \mathbf{C}(V_i) = 
    \begin{cases}
        1 & S_i \geq \mu - k \sigma \\
        0 & S_i < \mu - k \sigma
    \end{cases}
    \label{eq:outlier}
\end{equation} where $\mu$ represents the mean and $\sigma$ represents the standard deviation of average pairwise mutual information for each part, and k is a strictness constant. By this binary function, we automatically eliminate the outlier components to reach an ideal mix, where each piece is unique but also fits well with the others. Note that as the constant k decreases, the filter becomes more strict.

\subsection{Personalization of the Character}

Lastly, we train a LoRA \cite{hu2021lora} model with DreamBooth \cite{ruiz2022dreambooth} on the refined set of images in order to generate images across various contexts while maintaining the details of the character.

\section{Experiments}
\label{sec:experiments}

\begin{figure*}[]
    \centering
    \includegraphics[width=0.6\textwidth]{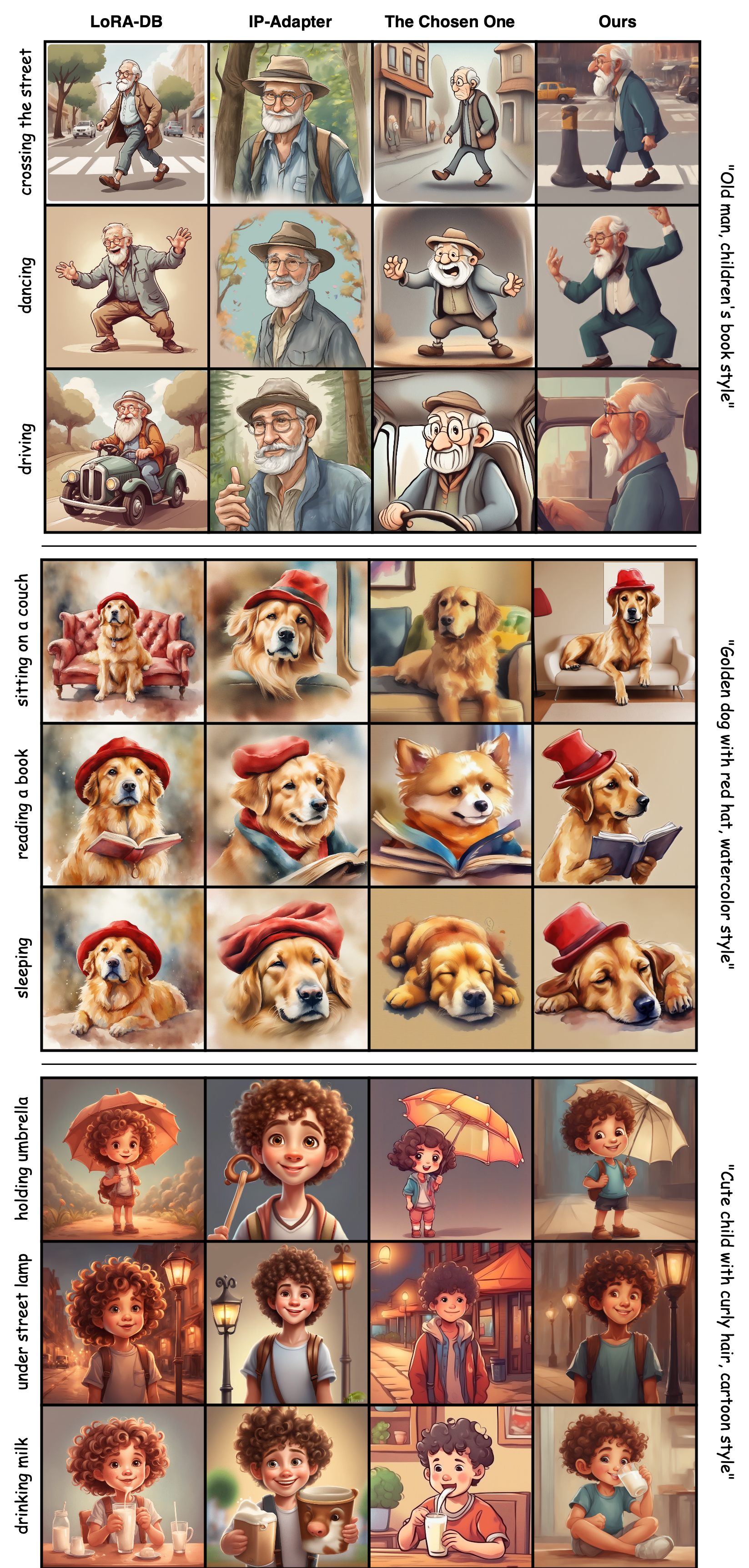}
    \caption{\textbf{Qualitative comparisons.} We compare ORACLE against various baselines, including LoRA DB, IP-Adapter, and The Chosen One. Our method surpasses these in effectively adhering to the given prompts and achieving greater consistency. }
    \label{fig:qual-comparison}
\end{figure*} 

We evaluate our method against various baselines through both quantitative and qualitative analysis. Following this, we detail the findings from our user study. Lastly, we demonstrate several applications of our approach, including story illustration, object generation, and 3D reconstruction.

\noindent \textbf{Baselines:} We compared our model with three  state-of-the-art models using  SDXL~\cite{podell2023sdxl}: The Chosen One~\cite{avrahami2023chosen}, IP-Adapter~\cite{ye2023ip-adapter}, and DreamBooth~\cite{ruiz2022dreambooth} combined with LoRA~\cite{hu2021lora} (LoRA DB). For each method, we used the same text prompt $\mathcal{P}$ such as \textit{'A golden dog with red hat, watercolor style'}. We use the same single image generated using prompt $\mathcal{P}$ for encoding-based models (IP-Adapter~\cite{ye2023ip-adapter}), or methods that require a reference image set (LoRA DB~\cite{ruiz2022dreambooth}). In our approach, we employ a refined set of images as outlined in our methodology.

\noindent \textbf{Implementation details:} We used the official codebases for all competitors. All methods use a recent state-of-the-art text-to-image model, SDXL~\cite{podell2023sdxl}. We run our experiments on a single A40 Nvidia GPU. Our method requires 30 seconds to generate the candidate set, and an additional 10 seconds to calculate mutual information for refinement. In contrast, methods such as \cite{avrahami2023chosen} takes 20 minutes per iteration. We used the strictness constant $k$ in Eq. \ref{eq:outlier} for outlier elimination as 1.

\subsection{Qualitative Experiments}
In Figure~\ref{fig:qual-result}, we showcase the qualitative outcomes of our methodology. Our approach effectively produces characters across diverse contexts and styles while preserving their identity. For instance, with the text prompt `witch, children's book style,' our method not only generates a distinct character but also adeptly positions it in various scenarios such as `on a beach,' `visiting the Colosseum,' `exploring the pyramids,' along with different activities like `writing a letter,' `listening to music,' `watching a butterfly,' or `playing ball.' Furthermore, our results demonstrate the capability to adapt these scenarios for a wide array of characters, including `an old woman,' `an elf,' `a bulldog,' `a pink owl,' and even photorealistic figures like `a woman with a purple scarf.'

\noindent \textbf{Qualitative comparison:} In Figure~\ref{fig:qual-comparison}, we provide a qualitative comparison between our method and other approaches. The IP-Adapter shows difficulty in adhering to prompts, such as \textit{"old man driving"} or \textit{"cute child with curly hair holding an umbrella"}. On the other hand, The Chosen One succeeds creating characters within specified concepts but faces challenges in generating characters that are consistent with the given prompt, as seen in the example of the \textit{`golden dog with red hat."} LoRA-DB~\cite{ruiz2022dreambooth} generally succeeds in producing consistent characters, yet it  fails to accurately follow the prompt, like in the case of the \textit{`golden dog with red hat sleeping."} Additionally, LoRA-DB~\cite{ruiz2022dreambooth} and IP-Adapter~\cite{ye2023ip-adapter} tend to keep the pose of characters unchanged across different contexts. In contrast, our method demonstrates superior ability in both following the prompt accurately and maintaining the character consistency.


\subsection{Quantitative Experiments}
We perform a quantitative evaluation based on two metrics: image prompt similarity and identity consistency. These metrics are widely used in studies on personalization techniques~\cite{ruiz2022dreambooth,gal2022textual} and generating consistent characters~\cite{avrahami2023chosen,tewel2024trainingfree}.  We measure the normalized cosine similarity between the image and prompt text embedding  using CLIP~\cite{radford2021learning} in order to evaluate the image prompt similarity. Similarly, we utilized CLIP to  assess identity consistency, where we calculate the average pairwise normalized cosine similarity among the images of the same subject among different contexts. We generate 4 characters, each in 12 different contexts, using the same seeds for each method, resulting in a total of 48 images evaluated for each method.
 
Our quantitative findings are illustrated in Figure~\ref{fig:quant-result}. Usually, finding a balance between preserving identity consistency and creating images that closely match the text prompt is essential. Techniques like LoRA DB~\cite{ruiz2022dreambooth} and IP-Adapter~\cite{ye2023ip-adapter} perform well in preserving character identity, primarily by performing minimal changes across images, but they tend to fall short of generating images that closely follow the text prompts. On the other hand, The Chosen One~\cite{avrahami2023chosen} is adept at creating images that match the prompts but struggle with keeping the character's identity consistent. Our method, however, achieves an optimal balance, effectively adhering to the text prompt while ensuring the character's identity remains consistent. This quantitative analysis corroborates our qualitative observations, underscoring the effectiveness of our approach.

\subsection{User Study}
We carried out a user study with 54 participants via the Prolific platform. Utilizing the visuals displayed in Fig. \ref{fig:qual-comparison}, we presented a series of three images for each character, depicted in different contexts. Participants were randomly assigned a set of images and instructed to rate them on a scale from 1 to 5, evaluating both their relevance to the text prompt (image-prompt similarity) and their consistency with each other (image-image similarity). Specifically, participants were asked the following questions:

\textit{Q1: Given the text description and the images shown above, how well the images reflect the given text description?  Rate from 1 (Not relevant at all) to 5 (Very Relevant)}

\textit{Q2: Considering the three images presented earlier, how consistent is the character depicted across them?  Rate 1 (Not consistent at all) to 5 (Very Consistent)}

The average ratings for all subjects are calculated for each method and are displayed in Figure~\ref{fig:user-study}. Overall, the results of the user study supports the quantitative findings presented in Figure~\ref{fig:quant-result}. Notably, our method emerged as the most preferred approach among users for both its consistency and relevance to the given text prompts. 

\begin{figure}[t]
    \includegraphics[width=0.46\textwidth]{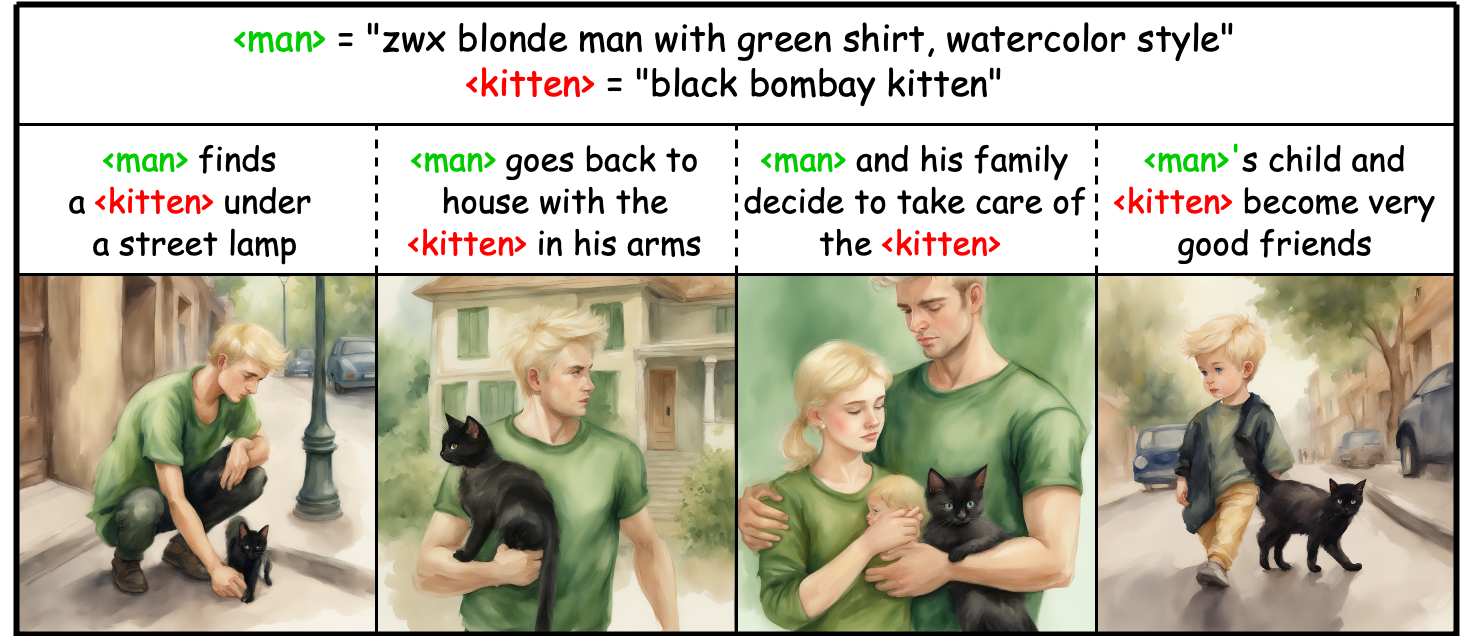}
    \caption{\textbf{Story illustration.} A demonstration of story illustration using a model trained with the specified description of the man.}
    \label{fig:story}
\end{figure}

\begin{figure}[h]
    \includegraphics[width=0.46\textwidth]{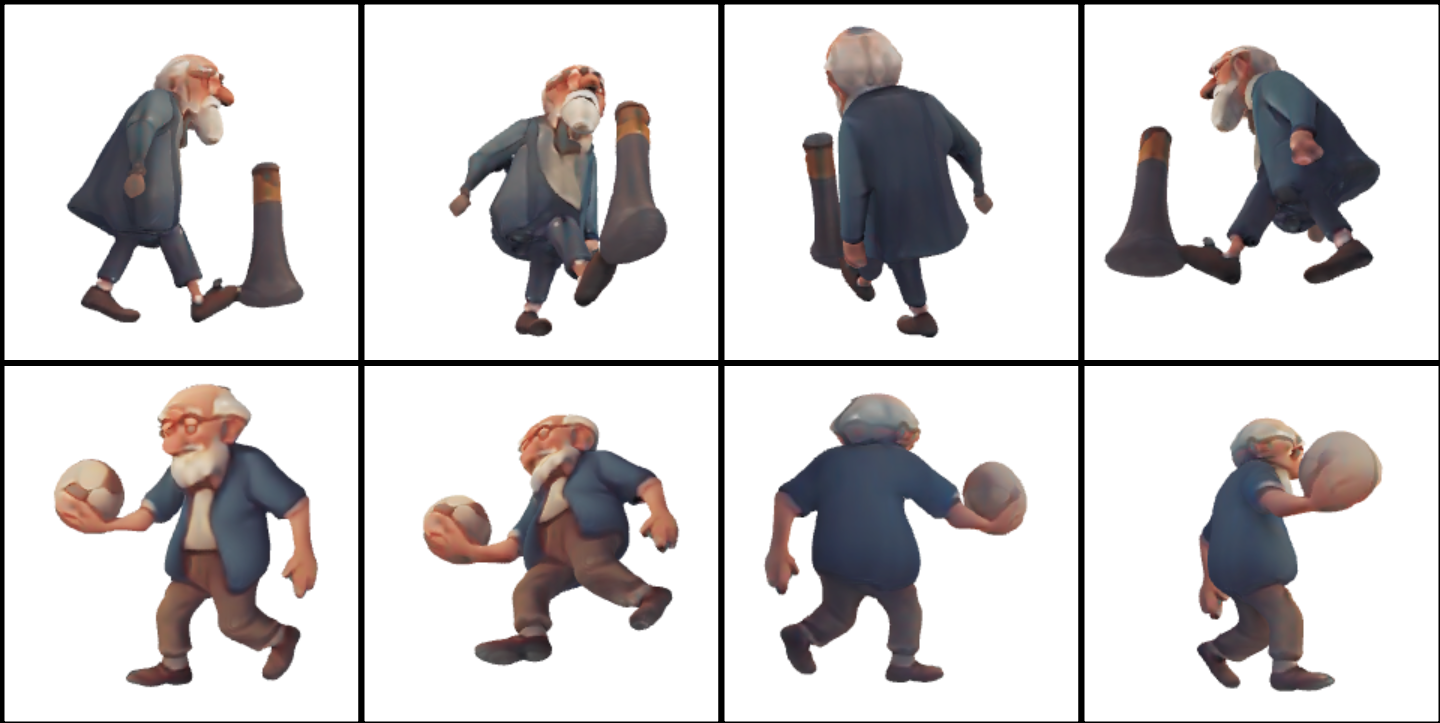}
    \caption{\textbf{An example of 3D character.}}
    \label{fig:3d-gen}
\end{figure}
 
\begin{figure}[h]
    \includegraphics[width=0.46\textwidth]{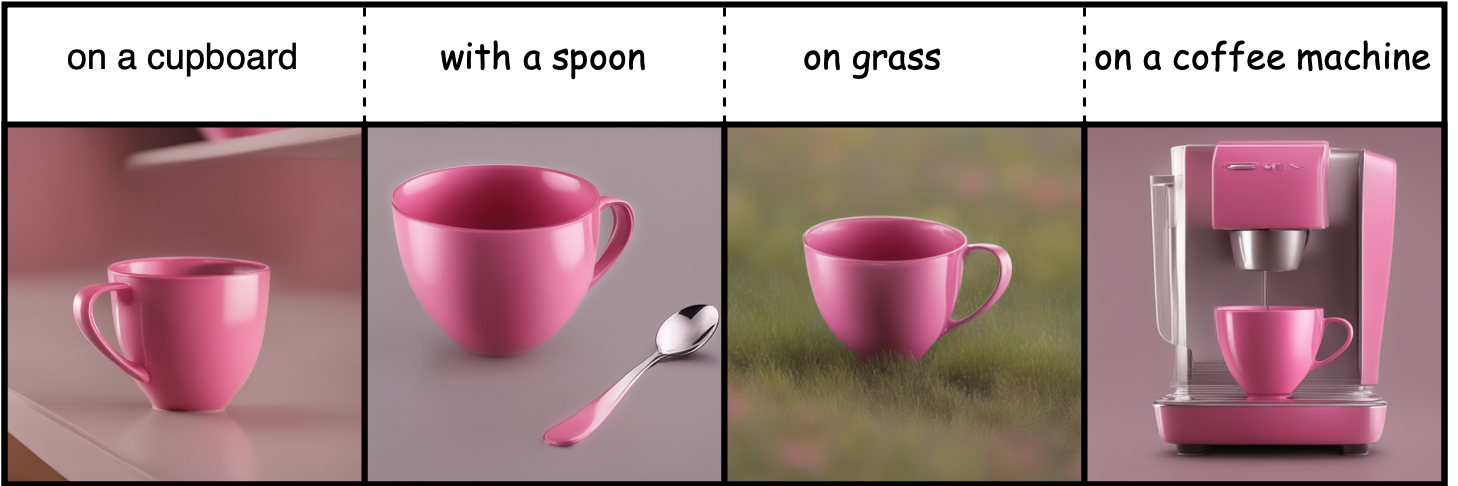}
    \caption{\textbf{An example of object generation.}}
    \label{fig:object}
\end{figure}
\subsection{Applications}
Our method have various applications, including story illustration, as demonstrated in Figure~\ref{fig:story}. Although the model is specifically trained for the man, it possesses the capability to generate images of the character's family or child, as illustrated in Figure~\ref{fig:story}. This capability  broadens the scope of illustrations achievable with a single trained model. Moreover, our characters can be transformed into 3D as in  Fig.~\ref{fig:3d-gen}, using off the shelf methods such as TripoSR~\cite{tochilkin2024triposr}.  Additionally, we illustrate that our method is also effective at generating consistent objects in addition to characters, as shown in Figure~\ref{fig:object}.

\section{Limitations}
\label{sec:limitations}

Even though our method can successfully generate consistent characters, inherent limitations associated with Stable Diffusion model exists; even if the images fed into the personalized model are perfectly consistent, the model might still alter certain details, such as clothing, across different contexts—a variation that might be desirable in specific scenarios.


\section{Conclusion}
\label{conc}

In conclusion, our work introduces a lightweight, fast, and efficient strategy for creating consistent characters through text-to-image models. Our experiments reveal that this approach successfully ensures consistency across various outputs and maintains alignment with the given text prompts, as evidenced by quantitative scores. This achievement is further validated by qualitative evaluations and a user study. Our method is opening up new avenues for utilizing text-to-image diffusion models to craft cohesive and captivating visual stories.

\bibliographystyle{iccc}
\bibliography{iccc}

\newpage

\onecolumn
\section{Appendix}
\label{sec:appendix}

\begin{figure*}[h]
    \centering
    \onecolumn \includegraphics[width=0.8\textwidth]{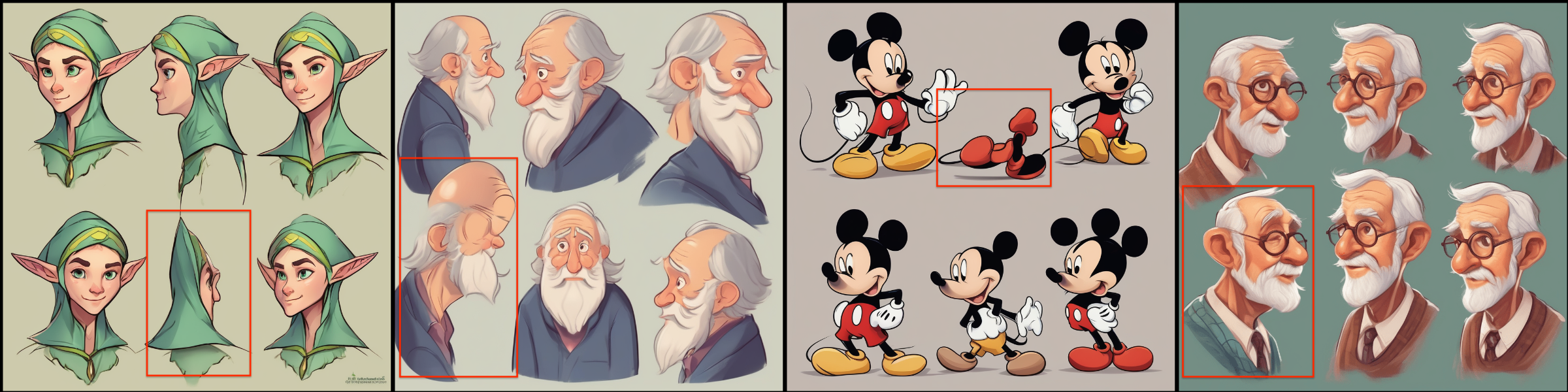}
    \caption{Examples of candidate set refinement using mutual information. Eliminated parts are indicated by red bounding boxes.}
    \label{fig:appendix}
\end{figure*}

\vspace{0.5cm}
We provide examples from candidate set refinement through mutual information in Figure \ref{fig:appendix}. Despite the capability of text-to-image models to generate grid images, certain parts may exhibit inconsistencies with others. As depicted in Figure \ref{fig:appendix}, our approach efficiently identifies and addresses these inconsistencies.

\end{document}